# Using eigenvectors of the bigram graph to infer morpheme identity

Mikhail Belkin     John Goldsmith

Department of Mathematics Department of Linguistics
University of Chicago
Chicago IL 60637
misha@math.uchicago.edu  ja-goldsmith@uchicago.edu

## Abstract

This paper describes the results of some experiments exploring statistical methods to infer syntactic categories from a raw corpus in an unsupervised fashion. It shares certain points in common with Brown et at (1992) and work that has grown out of that: it employs statistical techniques to derive categories based on what words occur adjacent to a given word. However, we use an eigenvector decomposition of a nearest-neighbor graph to produce a two-dimensional rendering of the words of a corpus in which words of the same syntactic category tend to form clusters and neighborhoods. We exploit this technique for extending the value of automatic learning of morphology. In particular, we look at the suffixes derived from a corpus by unsupervised learning of morphology, and we ask which of these suffixes have a consistent syntactic function (e.g., in English, *-ed* is primarily a mark of verbal past tense, does but *–s* marks both noun plurals and 3$^{rd}$ person present on verbs).

## 1   Introduction

This paper describes some results of our efforts to develop statistical techniques for unsupervised learning of syntactic word-behavior, with two specific goals: (1) the development of visualization tools displaying syntactic behavior of words, and (2) the development of quantitative techniques to test whether a given candidate set of words acts in a syntactically uniform way, in a given corpus.[1]

In practical terms, this means the development of computational techniques which accept a corpus in an unknown language as input, and produce as output a two-dimensional image, with each word identified as a point on the image, in such a fashion that words with similar syntactic behavior will be placed near to each other on the image.

We approach the problem in two stages: first, a nearest-neighbor analysis, in which a graph is constructed which links words whose distribution is similar, and second, what we might call a planar projection of this graph onto $R^2$, that is to say, a two-dimensional region, which is maximally faithful to the relations expressed by the nearest-neighbor graph.

## 2   Method

The construction of the nearest-neighbor graph is a process which allows for many linguistic and practical choices. Some of these we have experimented with, and others we have not, simply using parameter values that seemed to us to be reasonable.  Our goal is to develop a graph in which vertices represent words, and edges represent pairs of words whose distribution in a corpus is similar. We then develop a representation of the graph by a symmetric matrix, and compute a small number of the eigenvectors of the normalized laplacian for which the eigenvalues are smallest. These eigenvectors provide us with the coordinates necessary for our desired planar representation, as explained in section 2.2.

---

[1] We are grateful to Yali Amit for drawing our attention to Shi and Malik 1997, to Partha Niyogi for helpful comments throughout the development of this material, and to Jessie Pinkham for suggestions on an earlier draft of this paper.

## 2.1 Nearest-neighbor graph construction.

We begin with the reasonable working assumption that to determine the syntactic category of a given word *w*, it is the set of words which appears immediate before *w*, and the set of words that appears immediately after *w*, that gives the best immediate evidence of a word's syntactic behavior. In a natural sense, under that assumption, an explicit description of the behavior of a word *w* in a corpus is a sparse vector L = [$l_1$, $l_2$, …, $l_V$], of length V (where "V" is the number of words in the vocabulary of the corpus), indicating by $l_i$ how often each word $v_i$ occurs immediately to the left of *w*, and also an similar vector R, also of length V, indicating how often each word occurs immediately to the right of *w*. Paraphrasing this, we may view the syntactic behavior of a word in a corpus as being expressed by its location in a space of 2V dimensions, or a vector from the origin to this location; this space has a natural decomposition into two spaces, called Left and Right, each of dimension V.

Needless to say, such a representation is not directly illuminating -- nor does it provide a way to cogently present similarities or clusterings among words. We now construct a symmetrical graph ("LeftGraph"), whose vertices are the K most frequent words in the corpus. (We have experimented with K = 500 and K = 1000). For each word w, we compute the cosine of the angle between the vector **w** and the K-1 other words $w_i$: $\frac{w \cdot w_i}{|w||w_i|}$, and use this figure to select the N words closest to *w*. We have experimented with N = 5,10,20 and 50. We insert an edge ($v_i$, $v_j$) in LeftGraph if $v_i$ is one of the N words closest to $v_j$ or $v_j$ is one of the N words closest to $v_i$. We follow the same construction for RightGraph in the parallel fashion. In much of the discussion that follows, the reader may take whatever we say about LeftGraph to hold equally true of RightGraph when not otherwise stated.

## 2.2 Projection of nearest-neighbor graph by spectral decomposition

In the canonical matrix representation of a (unweighted) graph, an entry M(i,j), with i distinct from j, is 1 if the graph includes an edge (i,j) and 0 otherwise. All diagonal elements are zero. The *degree* of a vertex of a graph is the number of edges adjacent to it; the degree of the $m^{th}$ vertex, $d(v_m)$ is thus the sum of the values in the $m^{th}$ row of M. If we define D as the diagonal matrix whose entry D(m,m) is $d(v_m)$, the degree of $v_m$, then the *laplacian* of the graph is defined as D – M. The *normalized laplacian* L is defined as $D^{½} ( D – M ) D^{½}$. The effect of normalization on the laplacian is to divide the weight of an entry M(i,j) that represents the edge between $v_i$ and $v_j$ by $\frac{1}{\sqrt{d(v_i)d(v_j)}}$, and to set the values of the diagonal elements to 1.[2]

The laplacian is a symmetric matrix which is known to be positive semi-definite (Chung 1997). Therefore all the eigenvalues of the laplacian are non-negative. We return to the space of our observations by premultiplying the eigenvectors by $D^{½}$. We will refer to these eigenvectors derived from LeftGraph (pre-multiplied by $D^{½}$) as {$L_0$, $L_1$, …} and those derived from RightGraph as {$R_0$, $R_1$, …}.

Now, $L_0$ (and $R_0$) are trivial (they merely express the frequency of the words in the corpus), but $L_1$ and $L_2$ provide us with very useful information. They each consist of a vector with one coordinate for each word among the K most frequent words in the corpus, and thus can be conceived of as a 1-dimensional representation of the vocabulary. In particular, $L_1$ is the 1-dimensional representation that optimally preserves the notion of locality described by the graph we have just constructed, and the choice of the top N eigenvectors provides a representation which optimally preserves the graph-locality in N-space. By virtue of being eigenvectors in the same eigenvector decomposition, $L_1$ and $L_2$ are orthogonal, but subject to that limitation, the projection to $R^2$ using the coordinates of $L_1$ and $L_2$ is the 2-dimensional representation that best

---

[2] Our attention was drawn to the relevance of the normalized laplacian by Shi and Malik 1997, who explore a problem in the domain of vision. We are indebted to Chung 1997 on spectral graph theory.

preserves the locality described by the graph in question (Chung 1997, Belkin and Niyogi 2002).

Thus, to the extent that the syntactic behavior of a word can be characterized by the set of its immediate right- and left-hand neighbors (which is, to be sure, a great simplification of syntactic reality), using the lowest-valued eigenvectors provides a good graphical representation of words, in the sense that words with similar left-hand neighbors will be close together in the representation derived from the LeftGraph (and similarly for RightGraph).

### 2.3 Choice of graphs

We explore below two types of projection to 2 dimensions: plotting the $1^{st}$ and $2^{nd}$ eigenvectors of LeftGraph (and RightGraph), and plotting the $1^{st}$ eigenvectors of LeftGraph and RightGraph against each other. In all of these cases, we have built a graph using the 20 nearest neighbors. In future work, we would like to look at varying the number of nearest neighbors that are linked to a given word. From manual inspection, one can see that in all cases, the nearest two or three words are very similar; but the depth of the nearest neighbor list that reflects words of truly similar behavior is, roughly, inversely proportional to the frequency of the word. This is not surprising, in the sense that higher frequency words tend to be grammatical words, and for such words there are fewer members of the same category.

### 2.4 English

Figure 1 illustrates the results of plotting the $1^{st}$ and $2^{nd}$ eigenvectors of LeftGraph based on the first 1,000,000 words of the Brown corpus, and using the 1,000 most frequent words and constructing a graph based on the 20 nearest neighbors. Figure 2 illustrates the results derived from the first two eigenvectors of RightGraph.

Figures 1 and 2 suggest natural clusterings, based both on density and on the extreme values of the coordinates. In Figure 1 (LeftGraph), the bottom corner consists primarily of non-finite verbs (*be, do, make*); the left corner of finite verbs (*was, had, has*); the right corner primarily of nouns (*world, way, system*); while the top shows little homogeneity, though it includes the prepositions. See Appendix 1 for details; the words given in the appendix are a complete list of the words in a neighborhood that includes the extreme tip of the representation. As we move away from the extremes, in some cases we find a less homogeneous distribution of categories, while in others we find local pockets of linguistic homogeneity: for example, regions containing names of cities, others containing names of countries or languages.

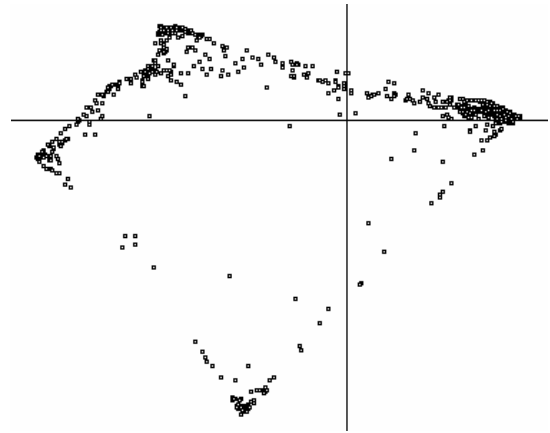

**Figure 1 English based on left-neighbors**

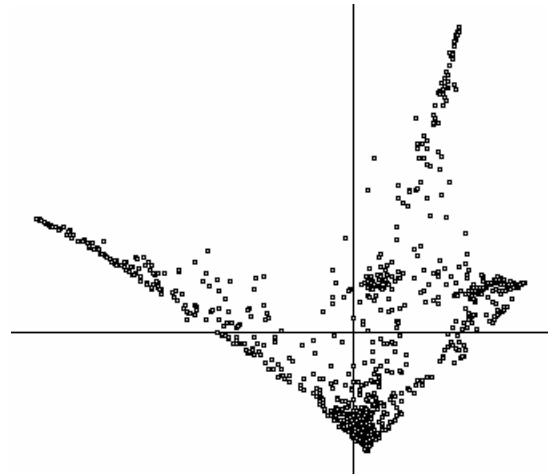

**Figure 2 English based on right-neighbors**

In Figure 2, the bottom corner consists of adjectives (*social, national, white*), the left corner of words that often are followed by *of* (*most, number, kind, secretary*), the right corner primarily by prepositions (*of, in for, on by*) and the top corner of words that often are followed by *to* (*going, wants, according*), (See Appendix 2 for details).

## 2.5 French

Figure 3 illustrates the results of plotting the 1st and 2nd eigenvectors of LeftGraph based on the first 1,000,000 words of a French encyclopedia, using the 1,000 most frequent words and constructing a graph based on the 20 nearest neighbors.

The bottom left tip of the figure consists entirely of feminine nouns (*guerre, population, fin*), the right tip of plural nouns (*années, états-unis, régions*), the top tip of finite verbs (*est, fut, a, avait*) plus *se* and *y*. A bit under the top tip one finds two sharp-tipped clusters; the one on the left consists of masculine nouns (*pays, sud, monde*). Other internal clusters, not surprisingly, are composed of words which, with high frequency, are preceded by a specific preposition (e.g., preceded by *à*: *peu, l'est, Paris*; by *en*: *particulier*, *effet*, and feminine names of geographical areas such as *France*).

Figure 4 illustrates plotting the 1st eigenvector of LeftGraph against the 1st eigenvector of RightGraph. We find a striking "striped" effect which is due to the masculine/feminine gender system of French. There are three stripes that stand out at the top of the figure. The one on the extreme left consists of singular feminine nouns, the one to its right, but left of center, consists of singular masculine nouns, and the one on the extreme right consists of plural nouns of both genders.

The lowest region of the graph, somewhat left of center, contains grammatical morphemes. At the very bottom are found relative and subordinating conjunctions (*où, car, lequel, laquel, lesquelles, lesquels, quand, si*), and just above them are the prepositions: *selon, durant, malgré, pendant, après, entre, jusqu'à, contre, sur*, etc.)

We find it striking that the gender system of French has such a pervasive impact upon the global form of the 1st eigenvector map as in Figure 4, and we plan to run further experiments with other language with gender systems to see the extent to which this result obtains consistently.

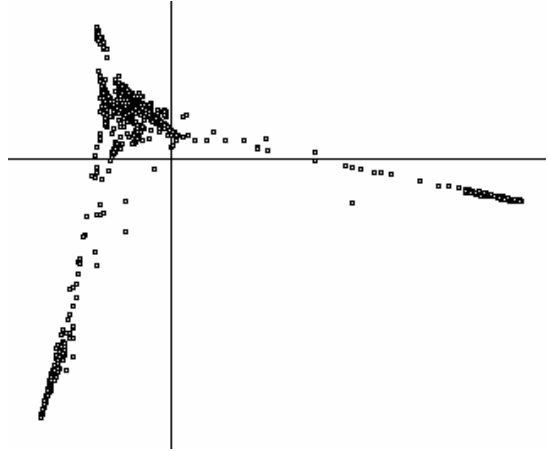

Figure 3 French based on left-neighbors

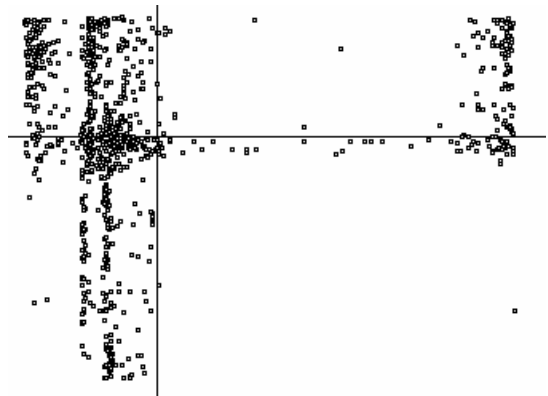

Figure 4 French 1st eigenvector of Left and Right

## 3 Identifying syntactic behavior of automatically identified suffixes

Interesting as they are, the representations we have seen are not capable of specifying membership in grammatical categories in an absolute sense. In this section, we explore the application of this representation to a text which has been morphologically analyzed by a language-neutral morphological analyzer. For this purpose, we employ the algorithm described in Goldsmith (2001), which takes an unanalyzed corpus and provides an analysis of the words into stems and suffixes. What is useful about that algorithm for our purposes is that it shares the same commitment to analysis based only on a raw (untreated) natural text, and neither hand-coding nor prior linguistic knowledge.

The algorithm in Goldsmith (2001) links each stem in the corpus to the set of suffixes (called its *signature*) with which it appears in the corpus. Thus the stem *jump* might appear with the three suffixes *ed-ing-s* in a given corpus.

But a morphological analyzer alone is not capable of determining whether the *–ed* that appears in the signature *ed-ing-s* is the same *–ed* suffix that appears in the signature *ed-ing* (for example), or whether the suffix *–s* in *ed-ing-s* is the same suffix that appears in the signature *NULL-s-'s* (this last signature is the one associated with the stem *boy* in a corpus containing the words *boy-boys-boy's*). A moment's reflection shows that the suffix *–ed* is indeed the same verbal past tense suffix in both cases, but the suffix *–s* is different: in the first case, it is a verbal suffix, while in the second it is a noun suffix.

In general, morphological information alone will not be able to settle these questions, and thus automatic morphology alone will not be able to determine which signatures should be "collapsed" (that is, *ed-ing-s* should be viewed as a special sub-case of the signature *NULL-ed-ing-s*, but *NULL-s* is not to be treated as a special case of *NULL-ed-ing-s*).

We therefore have asked whether the rudimentary syntactic analysis described in the present paper could provide the information needed for the automatic morphological analyzer.

The answer appears to be that if a suffix has an unambiguous syntactic function, then that suffix's identity can be detected automatically even when it appears in several different signatures. As we will see momentarily, the clear example of this is English *-ed*, which is (almost entirely) a verbal suffix. When a suffix is not syntactically homogeneous, then the words in which that suffix appears are scattered over a much larger region, and this difference appears to be quite sharply measurable.

### 3.1 The case of the verbal suffix –ed

In the automatic morphological analysis of the first 1,000,000 words of the Brown corpus that we produced, there are 26 signatures that contain the suffix *–ed*: *NULL.ed.s*, *e.ed.ing*, *NULL.ed.er.es.ing*, and 23 others of similar sort. We calculated a nearest neighbor graph as described above, with a slight variation. We considered the 1000 most frequent words to be atomic and unanalyzed morphologically, and then of the remaining words, we automatically replaced each *stem* with its corresponding signature. Thus as *jumped* is analyzed as *jump+ed*, and *jump* is assigned the signature *NULL.ed.er.s.ing* (based on the actual forms of the stem found in the corpus), the word *jumped* is replaced in the bigram calculations by the pseudo-word *NULL.ed.er.s.ing_ed*: the stem *jump* is replaced by its signature, and the actual suffix *-ed* remains unchanged, but is separated from its stem by an underscore _. Thus all words ending in *–ed* whose stems show the same *morphological* variations are treated as a single element, from the point of view of our present syntactic analysis.

We hoped, therefore, that these 26 signatures with *–ed* appended to them would appear very close to each other in our 2-dimensional representation, and this was exactly what we found.

To quantify this result, we calculated the coordinates of these 26 signatures in the following way. We normalize coordinates so that the lowest x-coordinate value is 0.0 and the highest is 1.0; likewise for the y-coordinates. Using these natural units, then, on the LeftGraph data, the average distance from each of the signatures to the center of these 26 points is 0.050. While we do not have at present a criterion to evaluate the closeness of this clustering, this appears to us at this point to be well within the range that an eventual criterion will establish. (A distance of 0.05 by this measure is a distance equal to 5% along either one of the axes, a fairly small distance.) On the RightGraph data, the average distance is 0.054.

### 3.2 The cases of –s and –ing

By contrast, when we look at the range of the 19 signatures that contain the suffix *–s*, the average distance to mean in the LeftGraph is 0.265, and in the RightGraph, 0.145; these points are much more widely scattered. We interpret this as being due to the fact that *–s* serves at least two functions: it marks the 3rd person present form of the verb, as well as the nominal plural.

Similarly, the suffix *–ing* marks both the verbal progressive form as well as the gerundive, used both as an adjective and as a noun, and we expect a scattering of these forms as a result. We find an average to mean of 0.096 in the LeftGraph, and of 0.143 in the RightGraph.

By way of even greater contrast, we can calculate the scatter of the NULL suffix, which is identified in all stems that appear without a suffix (e.g., the verb *play*, the noun *boy*). This "suffix" has an average distance to mean of 0.312 in the LeftGraph, and 0.192 in the RightGraph. This is the scatter we would expect of a group of words that have no linguistic coherence.

### 3.3 Additional suffixes tested

Suffix *–ly* occurs with five signatures, and an average distance to mean of 0.032 in LeftGraph, and 0.100 in RightGraph.[3] The suffix *'s* occurs in only two signatures, but their average distance to mean is 0.000 [sic] in LeftGraph, and 0.012 in RightGraph. Similarly, the suffix *–al* appears in two signatures (*NULL.al.s* and *NULL.al*), and their average distance to mean is 0.002 in LeftGraph, and also 0.002 in RightGraph. The suffix *–ate* appears in three signatures, with an average distance to mean of 0.069 in LeftGraph, and 0.080 in RightGraph. The suffix *–ment* appears in two signatures, with an average distance to mean of 0.012 in LeftGraph, and 0.009 in RightGraph.

### 3.4 French suffixes –ait, -er, -a, -ant, -e

We performed the same calculation for the French suffix *–ait* as for the English suffixes discussed above. *–ait* is the highest frequency 3rd person singular imperfect verbal suffix, and as such is one of the most common verbal suffixes, and it has no other syntactic functions. It appears in seven signatures composed of verbal suffixes, and they cluster well in the spaces of both LeftGraph and RightGraph, with an average distance to mean of 0.068 in the LeftGraph, and 0.034 in the RightGraph.

The French suffix *–er* is by far the most frequent infinitival marker, and it appears in 14 signatures, with an average distance to mean of 0.055 in LeftGraph, and 0.071 in RightGraph.

The 3rd singular simple past suffix *–a* appears in 11 signatures, and has an average distance to mean of 0.023 in LeftGraph, and 0.029 in RightGraph.

The present participle verbal suffix *–ant* appears in 10 suffixes, and has an average

---

[3] This latter figure deserves a bit more scrutiny; one of the five is an outlier: if we restricted our attention to four of them, the average distance to mean is 0.014.

distance to mean of 0.063 in LeftGraph, and of 0.088 in RightGraph.

On the other hand, the suffix *–e* appears as the last suffix in a syntactically heterogeneous set of words: nouns, verbs, and adjectives. It has an average distance to mean of 0.290 in LeftGraph and of 0.130 in RightGraph. This is as we expect: it is syntactically heterogeneous, and therefore shows a large average distance to mean.

### 3.5 Summary

Here are the average distances to mean for the cases where we expect syntactic coherence and the cases where we do not expect syntactic coherence. Our hypothesis is that the numbers will be small for the suffixes where we expect coherence, and large for those where we do not expect coherence, and this hypothesis is strongly borne out. We note empirically that we may take an average value of the two columns of .10 as a reasonable cut-off point.

| | **LeftGraph** | **RightGraph** |
|---|---|---|
| **Expect coherence:** | | |
| *ed* | 0.050 | 0.054 |
| *-ly* | 0.032 | 0.100 |
| *'s* | 0.000 | 0.012 |
| *-al* | 0.002 | 0.002 |
| *-ate* | 0.069 | 0.080 |
| *-ment* | 0.012 | 0.009 |
| *-ait* | 0.068 | 0.034 |
| *-er* | 0.055 | 0.071 |
| *-a* | 0.023 | 0.029 |
| *-ant* | 0.063 | 0.088 |
| | **LeftGraph** | **RightGraph** |
| **Expect little/no coherence:** | | |
| *-s* | 0.265 | 0.145 |
| *-ing* | 0.096 | 0.143 |
| *NULL* | 0.312 | 0.192 |
| *-e* | 0.290 | 0.130 |

**Figure 5 Average distance to mean of suffixes**

## 4 Conclusions

We have presented a simple yet mathematically sound method for representing the similarity of local syntactic behavior of words in a large corpus, and suggested one practical application. We have by no means exhausted the possibilities of this treatment. For example, it seems very reasonable to adjust the number of nearest neighbors permitted in the graph based on word-frequency: the higher the frequency, the fewer the number of nearest neighbors would be permitted in the graph. We leave this and other questions for future research.

This method does not appear strong enough at present to establish syntactic categories with sharp boundaries, but it is strong enough to determine with some reliability whether sets of words proposed by other, independent heuristics (such as presence of suffixes determined by unsupervised learning of morphology) are syntactically homogenous.

The reader can download the files discussed in this paper and a graphical viewer from http://humanities.uchicago.edu/faculty/goldsmith/eigenvectors/.

# Appendix 1

Typical examples from corners of Figure 1.
**Bottom:**

| be | do | me | make | see |
|---|---|---|---|---|
| get | take | go | say | put |
| find | give | provide | keep | run |
| tell | leave | pay | hold | live |

**Left:**

| was | had | has | would | said |
|---|---|---|---|---|
| could | did | might | went | thought |
| told | took | asked | knew | felt |
| began | saw | gave | looked | became |

**Right:**

| world | way | same | united | right |
|---|---|---|---|---|
| system | city | case | church | problem |
| company | past | field | cost | department |
| university | rate | center | door | surface |

**Top:**

| and | to | in | that | for |
|---|---|---|---|---|
| he | as | with | on | by |
| at | or | from | but | I |
| they | we | there | you | who |

# Appendix 2

Typical examples from corners of Figure 2.
**Bottom:**

| social | national | white | local | political |
|---|---|---|---|---|
| personal | private | strong | medical | final |
| black | French | technical | nuclear | british |
| health | husband | blue | | |

**Left:**

| most | number | kind | full | type |
|---|---|---|---|---|
| secretary | amount | front | instead | member |
| sort | series | rest | types | piece |
| image | lack | | | |

**Right:**

| of | in | for | on | by |
|---|---|---|---|---|
| at | from | into | after | through |
| under | since | during | against | among |
| within | along | across | including | near |

**Top:**

| going | want | seems | seemed | able |
|---|---|---|---|---|
| wanted | likely | difficult | according | due |
| tried | decided | trying | related | try |